\documentclass[11pt,a4paper]{article}
\usepackage[hyperref]{acl2021}
\usepackage{times}
\usepackage{latexsym}

\usepackage{amsmath}
\usepackage{amssymb}
\usepackage{graphicx}

\usepackage{microtype}

\aclfinalcopy 

\title{MLBiNet: A Cross-Sentence Collective Event Detection Network}

\author{
Dongfang Lou\textsuperscript{\rm 1,2 \thanks{ \quad Equal contribution and shared co-first authorship.} },  
Zhilin Liao \textsuperscript{\rm 1,2 \footnotemark[1]}, 
\textbf{Shumin Deng}\textsuperscript{\rm 1,2}, 
\textbf{Ningyu Zhang}\textsuperscript{\rm 1,2 \footnotemark[2]}, 
\textbf{Huajun Chen}\textsuperscript{\rm 1,2 \thanks{\quad Corresponding author.}} \\
\textsuperscript{\rm 1} Zhejiang University \& AZFT Joint Lab for Knowledge Engine \\
\textsuperscript{\rm 2} Hangzhou Innovation Center, Zhejiang University  \\
\texttt{loudongfang2015@163.com, zhilinliao@yeah.net}\\
\texttt{\{231sm,zhangningyu,huajunsir\}@zju.edu.cn} \\

}

\date{}

\begin{document}
\maketitle
\begin{abstract}
We consider the problem of collectively detecting multiple events, particularly in cross-sentence settings. The key to dealing with the problem is to encode semantic information and model event inter-dependency at a document-level. In this paper, we reformulate it as a Seq2Seq task and propose a \textbf{M}ulti-\textbf{L}ayer \textbf{Bi}directional \textbf{Net}work (MLBiNet) to capture the document-level association of events and semantic information simultaneously. Specifically, a bidirectional decoder is firstly devised to model event inter-dependency within a sentence when decoding the event tag vector sequence. Secondly, an information aggregation module is employed to aggregate sentence-level semantic and event tag information. Finally, we stack multiple bidirectional decoders and feed cross-sentence information, forming a multi-layer bidirectional tagging architecture to iteratively propagate information across sentences. We show that our approach provides significant improvement in performance compared to the current state-of-the-art results\footnote{The code is available in \url{https://github.com/zjunlp/DocED}.}.
\end{abstract}

\section{Introduction}
\label{sec:intro}
Event detection (ED) is a crucial sub-task of event extraction, which aims to identify and classify event triggers. For instance, the document shown in Table \ref{tab:example_intro_attack}, which contains six sentences $\{s_1,\dots,s_6\}$, the ED system is required to identify four events: an \emph{Injure} event triggered by ``injuries", two \emph{Attack} events triggered by ``firing" and ``fight", and a \emph{Die} event triggered by ``death". 

Detecting event triggers from natural language text is a challenge task because of the following problems:
a). \textbf{Sentence-level contextual representation and document-level information aggregation} \cite{chen2018collective, zhao2018document,shen2020hierarchical}. In ACE 2005 corpus, the arguments of a single event instance may be scattered in multiple sentences \cite{zheng2019doc2edag,ebner2019multi}, which indicates that document-level information aggregation is critical for ED task. What's more, a word in different contexts would express different meanings and trigger different events.  For example, in Table \ref{tab:example_intro_attack}, ``firing" in $s_3$ means the action of firing guns (\emph{Attack} event) or forcing somebody to leave their job (\emph{End\_Position} event). To specify its event type, cross-sentence information should be considered. 
b). \textbf{Intra-sentence and inter-sentence event inter-dependency modeling} \cite{liao2010using,chen2018collective,liu2018jointly}. For $s_4$ in Table \ref{tab:example_intro_attack}, an \emph{Attack} event is triggered by ``fight", and a \emph{Die} event is triggered by ``death". This kind of event co-occurrence is common in ACE 2005 corpus, we investigated the dataset and found that about 44.4\% of the triggers appeared in this way. The cross-sentence event co-occurrence shown in $s_4$ and $s_3$ is also very common. Therefore, modeling the sentence-level and document-level event inter-dependency is crucial for jointly detecting multiple events.

\begin{table}
    \centering
    {\small{
    \begin{tabular}{|l|}
    \hline
         $s_1$: what a brave young woman \\
         \hline
         $s_2$: did you hear about the \textbf{injuries}[\emph{Injure}] she sustained\\
         \hline
         $s_3$: did you hear about the \textbf{firing}[\emph{Attack}] she did \\
         \hline
         $s_4$: she was going to \textbf{fight}[\emph{Attack}] to the \textbf{death}[\emph{Die}] \\
         \hline
         $s_5$: she was captured but she was one tough cookie\\
         \hline
         $s_6$: god bless here \\
         \hline
    \end{tabular}
    }}
    \caption{An example document in ACE 2005 corpus with cross-sentence semantic enhancement and event inter-dependency. Specifically, semantic information of $s_2$ provides latent information to enhance $s_3$, and \emph{Attack} event in $s_4$ also contributes to $s_3$.}
    \label{tab:example_intro_attack}
\end{table}

To address those issues, previous approaches \cite{chen2015event,nguyen2016joint,liu2018jointly,yan2019event,liu2019exploiting,zhang2019joint} mainly focused on sentence-level event detection, neglecting the document-level event inter-dependency and semantic information. Some studies \cite{chen2018collective,zhao2018document} tried to integrate semantic information across sentences via the attention mechanism. For the document-level event inter-dependency modeling, \citet{liao2010using} extended the features with event types to capture dependencies between different events in a document. Although great progress has been made in ED task due to recent advances in deep learning, there is still no unified framework to model the document-level semantic information and event inter-dependency. 

We try to analyze the ACE 2005 data to re-understand the challenges encountered in ED task. Firstly, we find that event detection is essentially a special Seq2Seq task, in which the source sequence is a given document or sentence, and the event tag sequence is target of task. Seq2Seq tasks can be effectively modeled via the RNN-based encoder-decoder framework, in which the encoder captures rich semantic information, while the decoder generates a sequence of target symbols with inter-dependency been captured. This separate encoder and decoder framework can correspondingly deal with the semantic aggregation and event inter-dependency modeling challenges in ED task. Secondly, for the propagation of cross-sentence information, we find that the relevant information is mainly stored in several neighboring sentences, while little is stored in distant sentences. For example, as shown in Table \ref{tab:example_intro_attack}, it seems that $s_2$ and $s_4$ contribute more to $s_3$ than $s_1$ and $s_5$.

In this paper, we propose a novel \textbf{M}ulti-\textbf{L}ayer \textbf{Bi}directional \textbf{Net}work (MLBiNet) for ED task. A bidirectional decoder layer is firstly devised to decode the event tag vector corresponding to each token with forward and backward event inter-dependency been captured. Then, the event-related information in the sentence is summarized through a sentence information aggregation module. Finally, the multiple bidirectional tagging layers stacking mechanism is proposed to propagate cross-sentence information between adjacent sentences, and capture long-range information as the increasing of layers. We conducted experimental studies on ACE 2005 corpus to demonstrate its benefits in  cross-sentence joint event detection. Our contributions are summarized as follows:

\begin{itemize}
    \item We propose a novel bidirectional decoder model to explicitly capture bidirectional event inter-dependency within a sentence, alleviating long-range forgetting problem of traditional tagging structure;
    
    \item We propose a model called MLBiNet to propagate semantic and event inter-dependency information across sentences and detect multiple events collectively; 
    
    \item We achieve the best performance ($F_1$ value) on ACE 2005 corpus, surpassing the state-of-the-art by $1.9$ points.
\end{itemize}

\section{Approach}
\label{sec:approach}

Generally, event detection on ACE 2005 corpus is treated as a classification problem, which is to determine whether it forms a part of an event trigger. Specifically,  for a given document $d = \{s_1,\dots,s_n\}$, where $s_i=\{w_{i,1},\dots,w_{i,n_i}\}$ denotes the $i$-th sentence containing $n_i$ tokens. We are required to predict the triggered event type sequence $y_{i}=\{y_{i,1},\dots,y_{i,n_i}\}$ based on contextual information of $d$. Without  ambiguity, we omit the subscript $i$.

\begin{figure*}
  \centering
  \includegraphics[width=10.0cm,height=6.7cm]{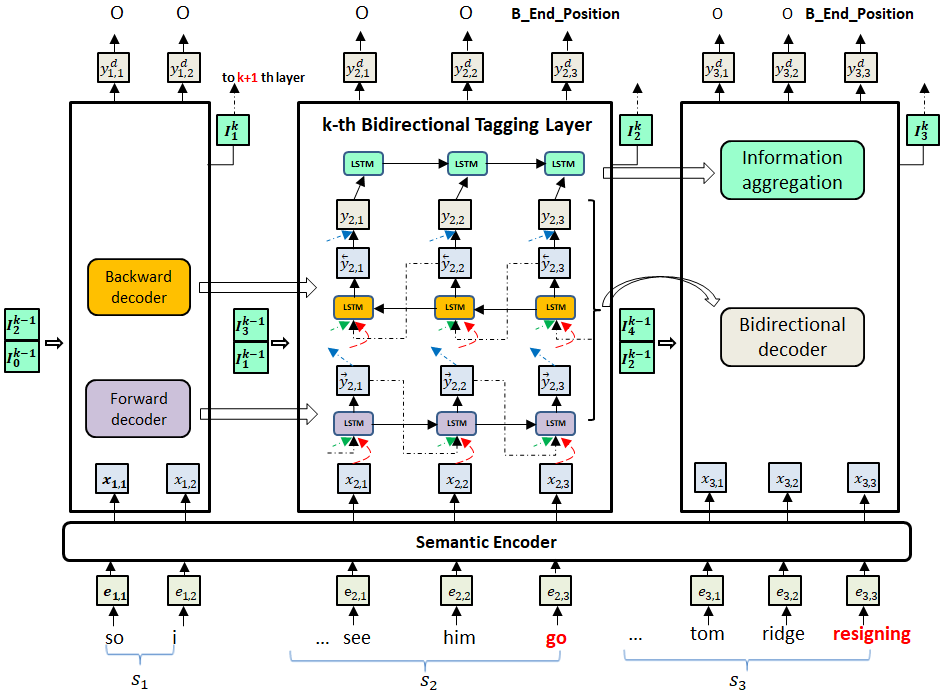}\\
  \caption{The architecture of our multi-layer bidirectional network (MLBiNet). The red arrow represents the input of semantic representation $\mathbf{x}_t$, the green arrow represents the input of adjacent sentences information $[\mathbf{I}_{i-1}^{k-1};\mathbf{I}_{i+1}^{k-1}]$ integrated in the previous layer, and the blue arrow represents the input of forward event tag vector. }
  \label{fig:model_structure}
\end{figure*}

For a given sentence, the event tags corresponding to tokens are associated, which is important for collectively detecting multiple events \cite{chen2018collective,liu2018jointly}. The way tokens are classified independently will miss the association. In order to capture the event inter-dependency, the sequential information of event tag should be retained. Intuitively, the ED task can be regarded as event tag sequence generation problem, which is essentially a Seq2Seq task. 
Specifically, the source sequence is a given document or sentence, and the event tag sequence to be generated is the target sequence. For instance, for sentence ``did you hear about the injuries she sustained", the decoder model is required to generate a tag sequence $[O, O, O, O, O, \mbox{\emph{B\_Injure}}, O, O]$, where ``O" denotes that the corresponding token is not part of event trigger and ``\emph{B\_Injure}" indicates an \emph{Injure} event is triggered.

We introduce the RNN-based encoder-decoder framework for ED task, considering that it is an efficient solution for Seq2Seq tasks. And we propose a multi-layer bidirectional network called MLBiNet shown in Figure \ref{fig:model_structure} to deal with the challenges in detecting multiple events collectively. The model framework consists of four components: the semantic encoder, the bidirectional decoder, the information aggregation module and stacking of multiple bidirectional tagging layers. We firstly introduce the encoder-decoder framework and discuss its compatibility with the ED task.

\subsection{Encoder–Decoder}\label{sec: encoder_decoder}
The RNN-based encoder-decoder framework \cite{cho2014learning,sutskever2014sequence,bahdanau2014neural,luong2015effective,gu2016incorporating} consists of two components: a) an encoder which converts the source sentence into a fixed length vector $\mathbf{c}$ and b) a decoder is to unfold the context vector $\mathbf{c}$ into the target sentence. As is formalized in \cite{gu2016incorporating}, the source sentence $s_i$ is converted into a fixed length vector $\mathbf{c}$ by the encoder RNN,  
\[
\textbf{h}_t = f(\textbf{h}_{t-1}, w_t), ~ \textbf{c} = \phi(\{\textbf{h}_1,\dots,\textbf{h}_{n_i}\})
\]
where $f$ is the RNN function, $\{\mathbf{h}_t\}$ are the RNN states, $w_t$ is the $t$-th token of source sentence, $\mathbf{c}$ is the so-called context vector, and $\phi$ summarizes the hidden states, e.g. choosing the last state $\mathbf{h}_{n_i}$. And the decoder RNN translates $\mathbf{c}$ into the target sentence according to:
\begin{equation}
\begin{gathered}
    \textbf{s}_t = f(y_{t-1},\textbf{s}_{t-1},\textbf{c}) \\
    p(y_t|y_{<t},s_i) = g(y_{t-1},\textbf{s}_t,\textbf{c})
\end{gathered}
\label{eqn:decoder}
\end{equation}
where $\mathbf{s}_t$ is the state at time $t$, $y_t$ is the predicted symbol at time $t$, $g$ is a classifier over the vocabulary, and $y_{<t}$ denotes the history $\{y_1,\dots,y_{t-1}\}$.

Studies \cite{bahdanau2014neural,luong2015effective} have shown that summarizing the entire source sentence into a fixed length vector will limit the performance of the decoder. They introduced the attention mechanism to dynamically changing context vector $\mathbf{c}_t$ in the decoding process, where $\mathbf{c}_t$ can be uniformly expressed as
\begin{gather}
    \mathbf{c}_t = \sum_{\tau = 1}^{n_i} \alpha_{t\tau}\mathbf{h}_\tau
    \label{eqn:ct}
\end{gather}
where $\alpha_{t\tau}$ is the contribution weight of $\tau$-th source token's state to context vector at time $t$, $\mathbf{h}_\tau$ denotes the representation of $\tau$-th token. 

We introduce the encoder-decoder framework to model ED task, mainly considering the following advantages: a) the separate encoder module is flexible in fusing sentence-level and document-level semantic information and b) the RNN decoder model \eqref{eqn:decoder} can capture sequential event tag dependency as the predicted tag vectors before $t$ will be used as input for predicting $t$-th symbol.

The encoder-decoder framework for ED task is slightly different from the general Seq2Seq task as follows: a) For ED task, the length of event tag sequence (target sequence) is known because its elements correspond one-to-one with tokens in the source sequence. However, the length of target sequence in the general Seq2Seq task is unknown. b) The vocabulary of decoder for ED task is a collection of event types, instead of words.


\subsection{Semantic Encoder} \label{sec: rnn_encoder}
In this module, we encode the sentence-level contextual information for each token with Bidirectional LSTM (BiLSTM) and self-attention mechanism. Firstly, each token is transformed into comprehensive representation by concatenating its word embedding and NER type embedding. The word embedding matrix is pretrained by Skip-gram model \cite{mikolov2013efficient}, and the NER type embedding matrix is randomly initialized and updated in the training process. For a given token $w_{t}$, its embedded vector is denoted as $\mathbf{e}_{t}$. 


We apply the BiLSTM \cite{zaremba2014learning} model for sentence-level semantic encoding, which can effectively capture sequential and contextual information for each token. The BiLSTM architecture is composed of a forward LSTM and a backward LSTM, i.e.,
$
        \overrightarrow{\mathbf{h}}_t = \overrightarrow{\mbox{LSTM}}(\overrightarrow{\mathbf{h}}_{t-1},\mathbf{e}_t), ~  \overleftarrow{\mathbf{h}}_t = \overleftarrow{\mbox{LSTM}}(\overleftarrow{\mathbf{h}}_{t+1},\mathbf{e}_t).
$
After encoding, the contextual representation of each token is $\mathbf{h}_t=[\overrightarrow{\mathbf{h}}_t;\overleftarrow{\mathbf{h}}_t]$. 

Attention mechanism between tokens within a sentence has been proven to further integrate long-range contextual semantic information.  For each token $w_t$, its contextual representation is the weighted average of the semantic information of all tokens in the sentence. We apply the attention mechanism proposed by \cite{luong2015effective} with the weights derived by 
\begin{equation}
    \begin{gathered}
    \alpha_{t,j} = \frac{\exp(z_{t,j})}{\sum_{m=1}^{n_i} \exp(z_{t,m})} \\
    z_{t,m} = \tanh(\mathbf{h}_t^\top W_{sa} \mathbf{h}_m +b_{sa})
    \end{gathered}
     \label{eqn:self-att-weights}
\end{equation}
And the contextual representation of $w_t$ is
$
    \mathbf{h}^a_t = \sum_{j=1}^{n_i} \alpha_{t,j} \mathbf{h}_j
$. By concatenating its lexical embedding and contextual representation, we get the final comprehensive semantic representation of $w_t$ as $\mathbf{x}_t = [\mathbf{h}^a_t;\mathbf{e}_t]$.

\subsection{Bidirectional Decoder}  \label{sec: rnn_decoder}
The decoder layer for ED task is to generate a sequence of event tags corresponding to tokens. As is noted, the tag sequence (target sequence) elements and tokens (source sequence) are in one-to-one correspondence. Therefore, the context vector $\mathbf{c}$ shown in \eqref{eqn:decoder} and \eqref{eqn:ct} can be personalized directly by $\mathbf{c}_t = \mathbf{x}_t$, which is equivalent to attention with degenerate weights. That is, $\alpha_{tt} = 1$ and $\alpha_{t\tau} = 0, ~\forall \tau \neq t$.

In traditional Seq2Seq tasks, the target sequence length is unknown during the inference process, so only the forward decoder is feasible. However, for the ED task, the length of the target sequence is known when given source sequence. Thus, we devise a bidirectional decoder to model event inter-dependency within a sentence. 

\paragraph{Forward Decoder} In addition to the semantic context vector $\mathbf{c}_t = \mathbf{x}_t$, the event information previously involved can help determine the event type triggered by $t$-th token. This kind of association can be captured by the forward decoder model:
\begin{equation}
\begin{gathered}
    \overset{\rightarrow}{\mathbf{s}}_t = f_{\mbox{fw}}(\overset{\rightarrow}{\mathbf{y}}_{t-1},\overset{\rightarrow}{\mathbf{s}}_{t-1},\mathbf{x}_t) \\
    \overset{\rightarrow}{\mathbf{y}}_t = \tilde{f}(W_y \overset{\rightarrow}{\mathbf{s}}_t + b_y)
\end{gathered}
\label{eqn:forward_decoder}
\end{equation}
where $f_{\mbox{fw}}$ is the forward RNN, $\{\overset{\rightarrow}{\mathbf{s}}_t\}$ are the states of forward RNN, $\{\overset{\rightarrow}{\mathbf{y}}_t\}$ are the forward event tag vectors. Compared with general decoder \eqref{eqn:decoder}, the classifier $g(\cdot)$ over vocabulary is replaced with a transformation $\tilde{f}(\cdot)$ (identity function, $\tanh$, sigmoid, etc.) to obtain the event tag vector.

\paragraph{Backward Decoder} Considering the associated events may also be mentioned later, we devise a backward decoder to capture this kind of dependency as follows:
\begin{equation}
\begin{gathered}
    \overset{\leftarrow}{\mathbf{s}}_t = f_{\mbox{bw}}(\overset{\leftarrow}{\mathbf{y}}_{t+1},\overset{\leftarrow}{\mathbf{s}}_{t+1},\mathbf{x}_t) \\
    \overset{\leftarrow}{\mathbf{y}}_t = \tilde{f}(W_y \overset{\leftarrow}{\mathbf{s}}_t + b_y)
\end{gathered}
\label{eqn:backward_decoder}
\end{equation}
where $f_{\mbox{bw}}$ is the backward RNN, $\{\overset{\leftarrow}{\mathbf{s}}_t\}$ are the states of backward RNN, $\{\overset{\leftarrow}{\mathbf{y}}_t\}$ are the backward event tag vectors.

\paragraph{Bidirectional Decoder} By concatenating $\overset{\rightarrow}{\mathbf{y}}_t$ and  $\overset{\leftarrow}{\mathbf{y}}_t$, we get the event tag vector $\mathbf{y}_t = [\overset{\rightarrow}{\mathbf{y}}_t; \overset{\leftarrow}{\mathbf{y}}_t]$ with bidirectional event inter-dependency been captured. The semantic and event-related entity information is also carried by $\mathbf{y}_t$ as $\mathbf{x}_t$ is an indirect input.

An alternative method modeling the sentence-level event inter-dependency called hierarchical tagging layer is proposed by \cite{chen2018collective}. The bidirectional decoder is quite different from the hierarchical tagging layer as follows:
\begin{itemize}
\item The bidirectional decoder models event inter-dependency immediately by combining a forward and a backward decoder. The hierarchical tagging layer utilizes two forward decoders and the tag attention mechanism to capture bidirectional event inter-dependency.

\item In the bidirectional decoder, the ED task is formalized as a special Seq2Seq task, which can simplify the event inter-dependency modeling problem and cross-sentence information propagation problem discussed below. 
\end{itemize}

The bidirectional RNN decoder unfolds the event tag vector corresponding to each token, and captures the bidirectional event inter-dependency within the sentence. To propagate information across sentences, we need to firstly aggregate useful information of each sentence.

\subsection{Information Aggregation} \label{sec: info_agg}
For current sentence $s_i$, the information we are concerned about can be summarized as recording which entities and tokens trigger which events. Thus, to summarize the information, we devise another LSTM layer (information aggregation module shown in Figure \ref{fig:model_structure}) with the event tag vector $\mathbf{y}_t$ as input. The information at $t$-th token is computed by 
\begin{gather}
    \tilde{\mathbf{I}}_t = \overrightarrow{\mbox{LSTM}}(\tilde{\mathbf{I}}_{t-1},\mathbf{y}_t) \label{eqn:info_agg}
\end{gather}
We choose the last state $\tilde{\mathbf{I}}_{n_i}$ as the summary information, which is $\mathbf{I}_{i} = \tilde{\mathbf{I}}_{n_i}$. 

The sentence-level information aggregation module bridges the information across sentences, as the well-formalized information can be easily integrated into the decoding process of other sentences, enhancing the event-related signal.

\subsection{Multi-Layer Bidirectional Network}  \label{sec: multi_layer_decoder}
In this module, we introduce a multiple bidirectional tagging layers stacking mechanism to aggregate information of adjacent sentences into the bidirectional decoder, and propagate information across sentences. The information $(\{\mathbf{y}_t\},\mathbf{I}_{i})$ obtained by the bidirectional decoder layer and information aggregation module has captured the event relevant information within a sentence. However, the cross-sentence information has not yet interacted. For a given sentence, as we can see in Table \ref{tab:example_intro_attack}, its relevant information is mainly stored in several neighboring sentences, while distant sentences are rarely relevant. Thus, we propose to transmit the summarized sentence information $\mathbf{I}_{i}$ among adjacent sentences. 

For the decoder framework shown in \eqref{eqn:forward_decoder} and \eqref{eqn:backward_decoder}, the cross-sentence information can be integrated by extending the input with $\mathbf{I}_{i-1}$ and $\mathbf{I}_{i+1}$. Further, we introduce a multiple bidirectional tagging layers stacking mechanism shown in Figure \ref{fig:model_structure} to iteratively aggregate information of adjacent sentences. The overall framework is named \textbf{M}ulti-\textbf{L}ayer \textbf{Bi}directional \textbf{Net}work (MLBiNet). As shown in Figure \ref{fig:model_structure}, a bidirectional tagging layer is composed of a bidirectional decoder and an information aggregation module. For sentence $s_i$, the outputs of $k$-th layer can be computed by
\begin{equation}
    \begin{gathered}
    \overset{\rightarrow}{\mathbf{s}}_t = f_{\mbox{fw}}(\overset{\rightarrow}{\mathbf{y}}_{t-1}^{k},\overset{\rightarrow}{\mathbf{s}}_{t-1},\mathbf{x}_t,\mathbf{I}_{i-1}^{k-1},\mathbf{I}_{i+1}^{k-1}) \\
    \overset{\leftarrow}{\mathbf{s}}_t = f_{\mbox{bw}}(\overset{\leftarrow}{\mathbf{y}}_{t+1}^{k},\overset{\leftarrow}{\mathbf{s}}_{t+1},\mathbf{x}_t,\mathbf{I}_{i-1}^{k-1},\mathbf{I}_{i+1}^{k-1}) \\
    \overset{\rightarrow}{\mathbf{y}}_t^{k} = \tilde{f}(W_y \overset{\rightarrow}{\mathbf{s}}_t + b_y) \\
    \overset{\leftarrow}{\mathbf{y}}_t^{k} = \tilde{f}(W_y \overset{\leftarrow}{\mathbf{s}}_t + b_y) \\
    \mathbf{y}_t^{k} = [\overset{\rightarrow}{\mathbf{y}}_t^{k};\overset{\leftarrow}{\mathbf{y}}_t^{k}]
    \end{gathered} 
    \label{eqn:cross_decoder}
\end{equation}
where $\mathbf{I}_{i-1}^{k-1}$ is the sentence information of $s_{i-1}$ aggregated in $\mbox{(k-1)}$-th layer, and $\{\mathbf{y}_t^{k}\}$ are event tag vectors obtained in $\mbox{k}$-th layer. The equation suggests that for each token of source sentence $s_i$, the input of cross-sentence information is identical $[\mathbf{I}_{i-1}^{k-1},\mathbf{I}_{i+1}^{k-1}]$. It is reasonable as their cross-sentence information available is the same for each token of current sentence. 

The iteration process shown in equation \eqref{eqn:cross_decoder} is actually an evolutionary diffusion of the cross-sentence semantic and event information in the document. Specifically, in the first tagging layer, information of current sentence is effectively modeled by the bidirectional decoder and information aggregation module. In the second layer, information of adjacent sentences is propagated to current sentence by plugging in $\mathbf{I}_{i-1}^{1}$ and $\mathbf{I}_{i+1}^{1}$ to the decoder. In general, in the $k$-th ($k\geq 3$) layer, since $s_{i-1}$ has captured the information of sentence $s_{i-k+1}$ in the $\mbox{(k-1)}$-th layer, then $s_i$ can obtain information in $s_{i-k+1}$ by acquiring the information in $s_{i-1}$. Thus, as the number of decoder layers increases, the model will capture information from distant sentences. For $K$-layer bidirectional tagging model, the sentence information with the longest distance of $\mbox{K-1}$ can be captured.


We define the final event tag vector of $w_t$ as the weighted sum of $\{\mathbf{y}_t^{k}\}_k$ in different layers, i.e.,
$
    \mathbf{y}^d_t = \sum_{k=1}^K \alpha^{k-1} \mathbf{y}_t^{k},
$
where $\alpha \in (0,1]$ is a weight decay parameter. It means that cross-sentence information can supplement to the current sentence, and the contribution gradually decreases as the distance increases when $\alpha < 1$.

We note that the parameters of bidirectional decoder and information aggregation module at different layers can be shared, because they encode and propagate the same structured information. In this paper, we set the parameters of different layers to be the same.

\subsection{Loss Function}
In order to train the networks, we minimize the negative log-likelihood loss function $J(\theta)$,
\begin{equation}
    J(\theta) = - \sum_{d\in D} \sum_{s\in d} \sum_{w_t\in s} \log p(O_t^{y_t}|d;\theta)
    \label{eqn:loss_fun}
\end{equation}
where $D$ denotes training documents set. The tag probability for token $w_t$ is computed by
\begin{equation}
    \begin{gathered}
        O_t = W_o \mathbf{y}^d_t + b_o \\
        p(O_t^j|d;\theta) = \exp(O^{j}_t) / \sum_{m=1}^{M} \exp(O_t^m) 
    \end{gathered}
    \label{eqn: output_layer}
\end{equation}
where $M$ is the number of event classes, $p(O_t^j|d;\theta)$ is the probability that assigning event type $j$ to token $w_t$ in document $d$ when parameter is $\theta$.

\section{Experiments}
\label{sec:experiments}
\subsection{Dataset and Settings}
We performed extensive experimental studies on the ACE 2005 corpus to demonstrate the effectiveness of our method on ED task. It defines 33 types of events and an extra \emph{NONE} type for the non-trigger tokens. We formalize it as a task to generate a sequence of 67-class event tag (with BIO tagging schema). The data splitting for training, validation and testing follows \cite{ji2008refining,chen2015event,liu2018jointly,chen2018collective,huang2020semi}, where the training set contains 529 documents, the validation set contains 30 documents and the remaining 40 documents are used as testing set.

We evaluated the performance of three multi-layer settings with 1-, 2- and 3-layer MLBiNet, respectively. We use the Adam \cite{kingma2017adam} for optimization. In all three settings, we cut every 8 consecutive sentences into a new document and padding when needed. Each sentence is truncated or padded to make it 50 in length. We set the dimension of word embedding as 100, the dimension of golden NER type and subtype embedding as 20. We set the dropout rate as $0.5$ and penalty coefficient as $2*10^{-5}$ to avoid overfitting. The hidden size of semantic encoder layer and decoder layer is set to 100 and 200, respectively. The size of forward and backward event tag vectors is set to 100. And we set the batch size as 64, the learning rate as $5*10^{-4}$ with decay rate 0.99, the weight decay parameter $\alpha$ as $1.0$. The results we report are the average of $10$ trials.

\subsection{Baselines}
For comparison, we investigated the performance of the following state-of-the-art methods: 1) \textbf{DMCNN} \cite{chen2015event}, which extracts multiple events from one sentence with dynamic multi-pooling CNN; 2) \textbf{HBTNGMA} \cite{chen2018collective}, which models sentence event inter-dependency via a hierarchical tagging model; 
3) \textbf{JMEE} \cite{liu2018jointly}, which models the sentence-level event inter-dependency via a graph model of the sentence syntactic parsing graph; 
4) \textbf{DMBERT-Boot} \cite{wang2019adversarial}, which augments the training data with external unlabeled data by adversarial mechanism; 5) \textbf{MOGANED} \cite{yan2019event}, which uses graph convolution network with aggregative attention to explicitly model and aggregate multi-order syntactic representations; 6) \textbf{SS-VQ-VAE} \cite{huang2020semi}, which learns to induct new event type by a semi-supervised vector quantized variational autoencoder framework, and fine-tunes with the pre-trained BERT-large model.

\subsection{Overall Performance}
{\small
\begin{table}
    \centering
    \begin{tabular}{|c|c|c|c|}
    \hline
       \textbf{Methods} & $\boldsymbol{P}$ & $\boldsymbol{R}$ & $\boldsymbol{F_1}$ \\
       \hline
      DMCNN  & 75.6 & 63.6 & 69.1 \\
      \hline
       HBTNGMA  & 77.9 & 69.1 & 73.3 \\
      \hline
      JMEE  & 76.3 & 71.3 & 73.7 \\
      \hline
      DMBERT-Boot  & 77.9 & 72.5 & 75.1 \\
       \hline
       MOGANED  & \textbf{79.5} & 72.3 & 75.7 \\
       \hline
       SS-VQ-VAE  & 75.7 & 77.8 & 76.7 \\
       \hline
       \hline
       \textbf{MLBiNet} (1-layer)  & 74.1 &  78.5 & 76.2 \\
           \hline
        \textbf{MLBiNet} (2-layer)  & 74.2 & \textbf{83.7} & \textbf{78.6} \\
           \hline
        \textbf{MLBiNet} (3-layer)   & 74.7 & 83.0 & \textbf{78.6}\\
        \hline
    \end{tabular}
    \caption{Performance comparison of different methods on
the test set with gold-standard entities.}
    \label{tab:exp_res}
\end{table}
}

Table \ref{tab:exp_res} presents the overall performance comparison between different methods with gold-standard entities. As shown, under 2-layer and 3-layer settings, our proposed model MLBiNet achieves better performance, surpassing the current state-of-the-art by $1.9$ points. More specifically, our models achieve higher recalls by at least $0.7$, $5.9$ and $5.2$ points, respectively.

The powerful encoder of BERT pre-trained model \cite{devlin2018bert} has been proven to improve the performance of downstream NLP tasks. The 2-layer MLBiNet outperforms BERT-Boot (BERT-base)  and SS-VQ-VAE (BERT-large) by $3.5$ and $1.9$ points, respectively. It proves the importance of event inter-dependency modeling and cross-sentence information integration for ED task. 

When only information of current sentence is available, the 1-layer MLBiNet outperforms HBTNGMA by $2.9$ points. It proves that the hierarchical tagging mechanism adopted by HBTNGMA is not as effective as the bidirectional decoding mechanism we proposed. Intuitively, the bidirectional decoder models event inter-dependency explicitly by a forward decoder and a backward decoder, which is more efficient than hierarchies.


\subsection{Effect on Extracting Multiple Events}

{\small
\begin{table}
    \centering
    \begin{tabular}{|c|c|c|c|}
    \hline
        \textbf{Methods} & \textbf{1/1} & \textbf{1/n} & \textbf{all} \\
         \hline
        DMCNN & 74.3 & 50.9 & 69.1 \\
         \hline
         HBTNGMA & 78.4 & 59.5 & 73.3 \\
          \hline
         JMEE & 75.2 & 72.7 & 73.7 \\
         \hline
         \hline
        \textbf{MLBiNet} (1-layer)  & 77.9 &  75.1 & 76.2 \\
           \hline
        \textbf{MLBiNet} (2-layer)  & \textbf{80.6} & 77.1 & \textbf{78.6} \\
           \hline
        \textbf{MLBiNet} (3-layer)  & 80.3 & \textbf{77.4} & \textbf{78.6}\\
        \hline
    \end{tabular}
    \caption{System Performance on Single Event Sentences (1/1) and Multiple Event Sentences (1/n). 1/1 means one sentence that has one event; otherwise, 1/n is used. ``\textbf{all}'' means all test data are included.}
    \label{tab:ablation_one_vs_more}
\end{table}
}

The existing event inter-dependency modeling methods \cite{chen2015event,chen2018collective,liu2018jointly} aim to extract multiple events jointly within a sentence. To demonstrate that sentence-level event inter-dependency modeling benefits from cross-sentence information propagation, we evaluated the performance of our model in single event extraction (1/1) and multiple events joint extraction (1/n). 1/1 means one sentence that has one event; otherwise, 1/n is used. 

The experimental results are presented in Table \ref{tab:ablation_one_vs_more}. As shown, we can verify the importance of cross-sentence information propagation mechanism and bidirectional decoder in sentence-level multiple events joint extraction based on the following results: a) When only the current sentence information is available, the 1-layer MLBiNet outperforms existing methods at least by $2.4$ points in 1/n case, which proves the effectiveness of bidirectional decoder we proposed; b) For ours 2-layer and 3-layer models, their performance in both 1/1 and 1/n cases surpasses the current methods by a large margin, which proves the importance of propagating information across sentences for single event and multiple events extraction. We conclude that it is the propagating information across sentences and bidirectional decoder which make cross-sentence joint event detection successful.

\subsection{Analysis of Decoder Layer}

{\small
\begin{table}
    \centering
    \begin{tabular}{|c|c|c|c|}
    \hline
        \textbf{Methods} & \textbf{1-layer} & \textbf{2-layer} & \textbf{3-layer} \\
         \hline
        backward & 72.2 & 75.0 & 75.5 \\
         \hline
         forward & 72.8 & 76.0 & 76.5 \\
          \hline
         \textbf{bidirectional} & \textbf{76.2} & \textbf{78.6} & \textbf{78.6} \\
         \hline
    \end{tabular}
    \caption{The performance of our proposed method with different multi-layer settings or decoder methods.}
    \label{tab:ablation_decoder_layer}
\end{table}
}

Table \ref{tab:ablation_decoder_layer} presents the performance of the model in three decoder mechanisms: forward, backward and bidirectional decoder, as well as three multi-layer settings. We can reach the following conclusions: a) Under three decoder mechanisms, the performance of the proposed model will be significantly improved as the number of decoder layers increases; b) The bidirectional decoder dominates both forward decoder and backward decoder, and forward decoder dominates backward decoder; c) The information propagation across sentences will enhance event relevant signal regardless of the decoder mechanism applied. Among the three decoder models, the bidirectional decoder performs best because of its ability in capturing bidirectional event inter-dependency, which proves both the forward and backward decoders are critical for event inter-dependency modeling.

\subsection{Analysis of Aggregation Model}

{\small
\begin{table}
    \centering
    \begin{tabular}{|c|c|c|c|}
    \hline
        \textbf{Methods} & \textbf{P} & \textbf{R} &  $\mathbf{F}_1$ \\
         \hline
        \emph{baseline} (1-layer) & 74.1 & 78.5 & 76.2 \\
        \hline
        \hline
        \emph{average} (2-layer) & 74.5 & 82.5 & 78.3 \\
         \hline
        \emph{concat} (2-layer) & \textbf{75.0} & 82.6 & \textbf{78.6} \\
          \hline
        \textbf{\emph{LSTM}} (2-layer) & 74.2 & \textbf{83.7} & \textbf{78.6} \\
         \hline
    \end{tabular}
    \caption{The performance of MLBiNet with different kinds of information aggregation mechanisms.}
    \label{tab:ablation_agg_method}
\end{table}
}

In information aggregation module, we introduce a \emph{LSTM} shown in \eqref{eqn:info_agg} to aggregate sentence information, and then propagate to other sentences via the bidirectional decoder. We compare other aggregation methods: a) \emph{concat} means the sentence information is aggregated by simply concatenating the first and last event tag vector of the sentence, and b) \emph{average} means the sentence information is aggregated by averaging the event tag vectors of tokens in the sentence. The experimental results are presented in Table \ref{tab:ablation_agg_method}.

Compared with the baseline 1-layer model, other three 2-layer settings equipped with information aggregation and cross-sentence propagation performs better. It proves that sentence information aggregation module can integrate some useful information and propagate it to other sentences through the decoder. On the other hand, the performance of \emph{LSTM} and \emph{concat} are comparable and stronger than \emph{average}. Considering that the input of the information aggregation module is the event tag vector obtained by the bidirectional decoder, which has captured the sequential event information. Therefore, it is not surprising that \emph{LSTM} does not have that great advantage over \emph{concat} and \emph{average}.


\section{Related Work}
Event detection is a well-studied task with research effort in the last decade. The existing methods \cite{chen2015event,nguyen2015event,liu2017exploiting,nguyen2018graph,deng2020meta,tong2020improving,lai2020event,liu2020event,li2020event,cui2020edge,ACL2021_OntoED,ACL2021_KEFSED} mainly focus on sentence-level event trigger extraction, neglecting the document information. Or the document-level semantic and event inter-dependency information are modeled separately. 

For the problem of event inter-dependency modeling, some methods were proposed to jointly extract triggers within a sentence. Among them, \citet{chen2015event} used dynamic multi-pooling CNN to preserve information of multiple events; \citet{nguyen2016joint} utilized the bidirectional recurrent neural networks to extract events; \citet{liu2018jointly} introduced syntactic shortcut arcs to enhance information flow and used graph neural networks to model graph information; \citet{chen2018collective} proposed a hierarchical tagging LSTM layer and tagging attention mechanism to model the event inter-dependency within a sentence. Considering that adjacent sentences also store some relevant event information, which would enhance the event signals of other sentences. These methods would miss the event inter-dependency information across sentences. For document-level event inter-dependency modeling, \citet{lin2020joint} proposed to incorporate global features to capture the cross-subtask and cross-instance interactions.

The deep learning methods on document-level semantic information aggregation are primarily based on multi-level attention mechanism. \citet{chen2018collective} integrated document information by introducing a multi-level attention. \citet{zhao2018document} used trigger and sentence supervised attention to aggregate information and enhance the sentence-level event detection. \citet{zheng2019doc2edag} utilized the memory network to store document level contextual information and entities. Some feature-based document level information aggregation methods were proposed by \cite{ji2008refining,liao2010using,hong2011using,huang2012modeling,reichart2012multi,lu2012automatic}. And \citet{zhang2020topic} proposed to aggregate the document-level information by latent topic modeling. The attention-based document-level information aggregation mechanisms treat all sentences in the document equally, which may introduce some noises from distant sentences. And the feature-based methods require extensive human engineering, which also greatly affects the portability of the model.

\section{Conclusions}
This paper presents a novel Multi-Layer Bidirectional Network (MLBiNet) to propagate document-level semantic and event inter-dependency information for event detection task. To the best of our knowledge, this is the first work to unify them in one model. Firstly, a bidirectional decoder is proposed to explicitly model the sentence-level event inter-dependency, and event relevant information within a sentence is aggregated by an information aggregation module. Then the multiple bidirectional tagging layers stacking mechanism is devised to iteratively propagate semantic and event-related information across sentence. We conducted extensive experiments on the widely-used ACE 2005 corpus, the results demonstrate the effectiveness of our model, as well as all modules we proposed. 

In the future, we will extend the model to the event argument extraction task and other information extraction tasks, where the document-level semantic aggregation and object inter-dependency are critical. For example, the recently concerned document-level relation extraction \cite{quirk2017distant, yao2019docred}, which requires reading multiple sentences in a document to extract entities and infer their relations by synthesizing all information of the document. For other sequence labeling tasks, such as the named entity recognition, we can also utilize the proposed architecture to model the entity label dependency.

\section*{Acknowledgments}
We  want to express gratitude to the anonymous reviewers for their hard work and kind comments. This work is funded by NSFCU19B2027/91846204, National Key R\&D Program of China (Funding No.SQ2018YFC000004). 

\bibliographystyle{acl_natbib}
\bibliography{anthology,acl2021}

\begin{thebibliography}{43}
\expandafter\ifx\csname natexlab\endcsname\relax\def\natexlab#1{#1}\fi

\bibitem[{Bahdanau et~al.(2015)Bahdanau, Cho, and Bengio}]{bahdanau2014neural}
Dzmitry Bahdanau, Kyunghyun Cho, and Yoshua Bengio. 2015.
\newblock Neural machine translation by jointly learning to align and
  translate.
\newblock \emph{CoRR}, abs/1409.0473.

\bibitem[{Chen et~al.(2015)Chen, Xu, Liu, Zeng, and Zhao}]{chen2015event}
Yubo Chen, Liheng Xu, Kang Liu, Daojian Zeng, and Jun Zhao. 2015.
\newblock Event extraction via dynamic multi-pooling convolutional neural
  networks.
\newblock In \emph{Proceedings of the 53rd Annual Meeting of the Association
  for Computational Linguistics and the 7th International Joint Conference on
  Natural Language Processing}, pages 167--176.

\bibitem[{Chen et~al.(2018)Chen, Yang, Liu, Zhao, and Jia}]{chen2018collective}
Yubo Chen, Hang Yang, Kang Liu, Jun Zhao, and Yantao Jia. 2018.
\newblock Collective event detection via a hierarchical and bias tagging
  networks with gated multi-level attention mechanisms.
\newblock In \emph{Proceedings of the 2018 Conference on Empirical Methods in
  Natural Language Processing}, pages 1267--1276.

\bibitem[{Cho et~al.(2014)Cho, Van~Merri{\"e}nboer, Gulcehre, Bahdanau,
  Bougares, Schwenk, and Bengio}]{cho2014learning}
Kyunghyun Cho, Bart Van~Merri{\"e}nboer, Caglar Gulcehre, Dzmitry Bahdanau,
  Fethi Bougares, Holger Schwenk, and Yoshua Bengio. 2014.
\newblock Learning phrase representations using rnn encoder-decoder for
  statistical machine translation.
\newblock \emph{Proceedings of the 2014 Conference on Empirical Methods in
  Natural Language Processing ({EMNLP})}.

\bibitem[{Cui et~al.(2020)Cui, Yu, Liu, Zhang, Wang, and Shi}]{cui2020edge}
Shiyao Cui, Bowen Yu, Tingwen Liu, Zhenyu Zhang, Xuebin Wang, and Jinqiao Shi.
  2020.
\newblock Edge-enhanced graph convolution networks for event detection with
  syntactic relation.
\newblock In \emph{Proceedings of the 2020 Conference on Empirical Methods in
  Natural Language Processing: Findings}, pages 2329--2339.

\bibitem[{Deng et~al.(2020)Deng, Zhang, Kang, Zhang, Zhang, and
  Chen}]{deng2020meta}
Shumin Deng, Ningyu Zhang, Jiaojian Kang, Yichi Zhang, Wei Zhang, and Huajun
  Chen. 2020.
\newblock Meta-learning with dynamic-memory-based prototypical network for
  few-shot event detection.
\newblock In \emph{Proceedings of the 13th International Conference on Web
  Search and Data Mining}, pages 151--159.

\bibitem[{Deng et~al.(2021)Deng, Zhang, Li, Chen, Tou, Chen, Huang, and
  Chen}]{ACL2021_OntoED}
Shumin Deng, Ningyu Zhang, Luoqiu Li, Hui Chen, Huaixiao Tou, Mosha Chen, Fei
  Huang, and Huajun Chen. 2021.
\newblock Ontoed: Low-resource event detection with ontology embedding.
\newblock In \emph{{ACL}}. Association for Computational Linguistics.

\bibitem[{Devlin et~al.(2018)Devlin, Chang, Lee, and
  Toutanova}]{devlin2018bert}
Jacob Devlin, Ming-Wei Chang, Kenton Lee, and Kristina Toutanova. 2018.
\newblock Bert: Pre-training of deep bidirectional transformers for language
  understanding.
\newblock \emph{Proceedings of the 2019 Conference of the North {A}merican
  Chapter of the Association for Computational Linguistics: Human Language
  Technologies, Volume 1 (Long and Short Papers)}.

\bibitem[{Ebner et~al.(2019)Ebner, Xia, Culkin, Rawlins, and
  Van~Durme}]{ebner2019multi}
Seth Ebner, Patrick Xia, Ryan Culkin, Kyle Rawlins, and Benjamin Van~Durme.
  2019.
\newblock Multi-sentence argument linking.
\newblock \emph{arXiv preprint arXiv:1911.03766}.

\bibitem[{Gu et~al.(2016)Gu, Lu, Li, and Li}]{gu2016incorporating}
Jiatao Gu, Zhengdong Lu, Hang Li, and Victor~O.K. Li. 2016.
\newblock Incorporating copying mechanism in sequence-to-sequence learning.
\newblock In \emph{Proceedings of the 54th Annual Meeting of the Association
  for Computational Linguistics (Volume 1: Long Papers)}, pages 1631--1640.

\bibitem[{Hong et~al.(2011)Hong, Zhang, Ma, Yao, Zhou, and Zhu}]{hong2011using}
Yu~Hong, Jianfeng Zhang, Bin Ma, Jianmin Yao, Guodong Zhou, and Qiaoming Zhu.
  2011.
\newblock Using cross-entity inference to improve event extraction.
\newblock In \emph{Proceedings of the 49th Annual Meeting of the Association
  for Computational Linguistics: Human Language Technologies}, pages
  1127--1136.

\bibitem[{Huang and Ji(2020)}]{huang2020semi}
Lifu Huang and Heng Ji. 2020.
\newblock Semi-supervised new event type induction and event detection.
\newblock In \emph{Proceedings of the 2020 Conference on Empirical Methods in
  Natural Language Processing (EMNLP)}, pages 718--724.

\bibitem[{Huang and Riloff(2012)}]{huang2012modeling}
Ruihong Huang and Ellen Riloff. 2012.
\newblock Modeling textual cohesion for event extraction.
\newblock \emph{Proceedings of the 26th Conference on Artificial Intelligence}.

\bibitem[{Ji and Grishman(2008)}]{ji2008refining}
Heng Ji and Ralph Grishman. 2008.
\newblock Refining event extraction through cross-document inference.
\newblock In \emph{Proceedings of ACL-08: HLT}, pages 254--262.

\bibitem[{Kingma and Ba(2017)}]{kingma2017adam}
Diederik~P. Kingma and Jimmy Ba. 2017.
\newblock \href {http://arxiv.org/abs/1412.6980} {Adam: A method for stochastic
  optimization}.

\bibitem[{Lai et~al.(2020)Lai, Nguyen, and Nguyen}]{lai2020event}
Viet~Dac Lai, Tuan~Ngo Nguyen, and Thien~Huu Nguyen. 2020.
\newblock Event detection: Gate diversity and syntactic importance scoresfor
  graph convolution neural networks.
\newblock \emph{arXiv preprint arXiv:2010.14123}.

\bibitem[{Li et~al.(2020)Li, Peng, Chen, Wang, Pan, Lyu, and Zhu}]{li2020event}
Fayuan Li, Weihua Peng, Yuguang Chen, Quan Wang, Lu~Pan, Yajuan Lyu, and Yong
  Zhu. 2020.
\newblock Event extraction as multi-turn question answering.
\newblock In \emph{Proceedings of the 2020 Conference on Empirical Methods in
  Natural Language Processing: Findings}, pages 829--838.

\bibitem[{Liao and Grishman(2010)}]{liao2010using}
Shasha Liao and Ralph Grishman. 2010.
\newblock Using document level cross-event inference to improve event
  extraction.
\newblock In \emph{Proceedings of the 48th Annual Meeting of the Association
  for Computational Linguistics}, pages 789--797.

\bibitem[{Lin et~al.(2020)Lin, Ji, Huang, and Wu}]{lin2020joint}
Ying Lin, Heng Ji, Fei Huang, and Lingfei Wu. 2020.
\newblock A joint neural model for information extraction with global features.
\newblock In \emph{Proceedings of the 58th Annual Meeting of the Association
  for Computational Linguistics}, pages 7999--8009.

\bibitem[{Liu et~al.(2019)Liu, Chen, and Liu}]{liu2019exploiting}
Jian Liu, Yubo Chen, and Kang Liu. 2019.
\newblock Exploiting the ground-truth: An adversarial imitation based knowledge
  distillation approach for event detection.
\newblock In \emph{Proceedings of the AAAI Conference on Artificial
  Intelligence}, volume~33, pages 6754--6761.

\bibitem[{Liu et~al.(2020)Liu, Chen, Liu, Bi, and Liu}]{liu2020event}
Jian Liu, Yubo Chen, Kang Liu, Wei Bi, and Xiaojiang Liu. 2020.
\newblock Event extraction as machine reading comprehension.
\newblock In \emph{Proceedings of the 2020 Conference on Empirical Methods in
  Natural Language Processing (EMNLP)}, pages 1641--1651.

\bibitem[{Liu et~al.(2017)Liu, Chen, Liu, and Zhao}]{liu2017exploiting}
Shulin Liu, Yubo Chen, Kang Liu, and Jun Zhao. 2017.
\newblock Exploiting argument information to improve event detection via
  supervised attention mechanisms.
\newblock In \emph{Meeting of the Association for Computational Linguistics},
  pages 1789--1798.

\bibitem[{Liu et~al.(2018)Liu, Luo, and Huang}]{liu2018jointly}
Xiao Liu, Zhunchen Luo, and Heyan Huang. 2018.
\newblock Jointly multiple events extraction via attention-based graph
  information aggregation.
\newblock \emph{Proceedings of the 2018 Conference on Empirical Methods in
  Natural Language Processing}.

\bibitem[{Lu and Roth(2012)}]{lu2012automatic}
Wei Lu and Dan Roth. 2012.
\newblock Automatic event extraction with structured preference modeling.
\newblock In \emph{Proceedings of the 50th Annual Meeting of the Association
  for Computational Linguistics}, pages 835--844.

\bibitem[{Luong et~al.(2015)Luong, Pham, and Manning}]{luong2015effective}
Thang Luong, Hieu Pham, and Christopher~D. Manning. 2015.
\newblock Effective approaches to attention-based neural machine translation.
\newblock In \emph{Proceedings of the 2015 Conference on Empirical Methods in
  Natural Language Processing}, pages 1412--1421.

\bibitem[{Mikolov et~al.(2013)Mikolov, Chen, Corrado, and
  Dean}]{mikolov2013efficient}
Tomas Mikolov, Kai Chen, Gregory~S. Corrado, and Jeffrey Dean. 2013.
\newblock Efficient estimation of word representations in vector space.
\newblock \emph{International Conference on Learning Representations},
  abs/1301.3781.

\bibitem[{Nguyen et~al.(2016)Nguyen, Cho, and Grishman}]{nguyen2016joint}
Thien~Huu Nguyen, Kyunghyun Cho, and Ralph Grishman. 2016.
\newblock Joint event extraction via recurrent neural networks.
\newblock In \emph{Proceedings of the 2016 Conference of the North American
  Chapter of the Association for Computational Linguistics: Human Language
  Technologies}, pages 300--309.

\bibitem[{Nguyen and Grishman(2015)}]{nguyen2015event}
Thien~Huu Nguyen and Ralph Grishman. 2015.
\newblock Event detection and domain adaptation with convolutional neural
  networks.
\newblock In \emph{Proceedings of the 53rd Annual Meeting of the Association
  for Computational Linguistics and the 7th International Joint Conference on
  Natural Language Processing}, pages 365--371.

\bibitem[{Nguyen and Grishman(2018)}]{nguyen2018graph}
Thien~Huu Nguyen and Ralph Grishman. 2018.
\newblock Graph convolutional networks with argument-aware pooling for event
  detection.
\newblock In \emph{Proceedings of the Thirty-Second {AAAI} Conference on
  Artificial Intelligence}, pages 5900--5907.

\bibitem[{Quirk and Poon(2017)}]{quirk2017distant}
Chris Quirk and Hoifung Poon. 2017.
\newblock Distant supervision for relation extraction beyond the sentence
  boundary.
\newblock \emph{Proceedings of the 15th Conference of the {E}uropean Chapter of
  the Association for Computational Linguistics}.

\bibitem[{Reichart and Barzilay(2012)}]{reichart2012multi}
Roi Reichart and Regina Barzilay. 2012.
\newblock Multi event extraction guided by global constraints.
\newblock In \emph{Proceedings of the 2012 Conference of the North American
  Chapter of the Association for Computational Linguistics: Human Language
  Technologies}, pages 70--79.

\bibitem[{Shen et~al.(2020)Shen, Qi, Li, Bi, and Wang}]{shen2020hierarchical}
Shirong Shen, Guilin Qi, Zhen Li, Sheng Bi, and Lusheng Wang. 2020.
\newblock Hierarchical chinese legal event extraction via pedal attention
  mechanism.
\newblock In \emph{Proceedings of the 28th International Conference on
  Computational Linguistics}, pages 100--113.

\bibitem[{Shen et~al.(2021)Shen, Wu, Qi, Li, Haffari, and Bi}]{ACL2021_KEFSED}
Shirong Shen, Tongtong Wu, Guilin Qi, Yuan-Fang Li, Gholamreza Haffari, and
  Sheng Bi. 2021.
\newblock Adaptive knowledge-enhanced bayesian meta-learning for few-shot event
  detection.
\newblock In \emph{Findings of {ACL}}. Association for Computational
  Linguistics.

\bibitem[{Sutskever et~al.(2014)Sutskever, Vinyals, and
  Le}]{sutskever2014sequence}
Ilya Sutskever, Oriol Vinyals, and Quoc~V Le. 2014.
\newblock Sequence to sequence learning with neural networks.
\newblock In \emph{Advances in neural information processing systems}, pages
  3104--3112.

\bibitem[{Tong et~al.(2020)Tong, Xu, Wang, Cao, Hou, Li, and
  Xie}]{tong2020improving}
Meihan Tong, Bin Xu, Shuai Wang, Yixin Cao, Lei Hou, Juanzi Li, and Jun Xie.
  2020.
\newblock Improving event detection via open-domain trigger knowledge.
\newblock In \emph{Proceedings of the 58th Annual Meeting of the Association
  for Computational Linguistics}, pages 5887--5897.

\bibitem[{Wang et~al.(2019)Wang, Han, Liu, Sun, and Li}]{wang2019adversarial}
Xiaozhi Wang, Xu~Han, Zhiyuan Liu, Maosong Sun, and Peng Li. 2019.
\newblock Adversarial training for weakly supervised event detection.
\newblock In \emph{Proceedings of the 2019 Conference of the North American
  Chapter of the Association for Computational Linguistics: Human Language
  Technologies, Volume 1 (Long and Short Papers)}, pages 998--1008.

\bibitem[{Yan et~al.(2019)Yan, Jin, Meng, Guo, and Cheng}]{yan2019event}
Haoran Yan, Xiaolong Jin, Xiangbin Meng, Jiafeng Guo, and Xueqi Cheng. 2019.
\newblock Event detection with multi-order graph convolution and aggregated
  attention.
\newblock In \emph{Proceedings of the 2019 Conference on Empirical Methods in
  Natural Language Processing and the 9th International Joint Conference on
  Natural Language Processing (EMNLP-IJCNLP)}, pages 5770--5774.

\bibitem[{Yao et~al.(2019)Yao, Ye, Li, Han, Lin, Liu, Liu, Huang, Zhou, and
  Sun}]{yao2019docred}
Yuan Yao, Deming Ye, Peng Li, Xu~Han, Yankai Lin, Zhenghao Liu, Zhiyuan Liu,
  Lixin Huang, Jie Zhou, and Maosong Sun. 2019.
\newblock Docred: A large-scale document-level relation extraction dataset.
\newblock \emph{Proceedings of the 57th Annual Meeting of the Association for
  Computational Linguistics}.

\bibitem[{Zaremba and Sutskever(2014)}]{zaremba2014learning}
Wojciech Zaremba and Ilya Sutskever. 2014.
\newblock Learning to execute.
\newblock \emph{arXiv preprint arXiv:1410.4615}.

\bibitem[{Zhang et~al.(2020)Zhang, Liu, and Zhang}]{zhang2020topic}
Junchi Zhang, Mengchi Liu, and Yue Zhang. 2020.
\newblock Topic-informed neural approach for biomedical event extraction.
\newblock \emph{Artificial Intelligence in Medicine}, 103:101783.

\bibitem[{Zhang et~al.(2019)Zhang, Ji, and Sil}]{zhang2019joint}
Tongtao Zhang, Heng Ji, and Avirup Sil. 2019.
\newblock Joint entity and event extraction with generative adversarial
  imitation learning.
\newblock \emph{Data Intelligence}, 1(2):99--120.

\bibitem[{Zhao et~al.(2018)Zhao, Jin, Wang, and Cheng}]{zhao2018document}
Yue Zhao, Xiaolong Jin, Yuanzhuo Wang, and Xueqi Cheng. 2018.
\newblock Document embedding enhanced event detection with hierarchical and
  supervised attention.
\newblock In \emph{Proceedings of the 56th Annual Meeting of the Association
  for Computational Linguistics}, pages 414--419.

\bibitem[{Zheng et~al.(2019)Zheng, Cao, Xu, and Bian}]{zheng2019doc2edag}
Shun Zheng, Wei Cao, Wei Xu, and Jiang Bian. 2019.
\newblock {D}oc2{EDAG}: An end-to-end document-level framework for {C}hinese
  financial event extraction.
\newblock In \emph{Proceedings of the 2019 Conference on Empirical Methods in
  Natural Language Processing and the 9th International Joint Conference on
  Natural Language Processing (EMNLP-IJCNLP)}, pages 337--346.

\end{thebibliography}


\end{document}